\documentclass[runningheads]{llncs}
\usepackage[T1]{fontenc}
\usepackage{graphicx}
\usepackage{amsmath}
\usepackage{bbm}
\usepackage{algorithm}
\usepackage{algorithmic}

\usepackage{booktabs}

\begin{document}
\title{Evaluating Accuracy and Probabilistic Reliability of Zero-Shot Time Series Foundation Models
}

%
%
\author{
Panagiotis Michael\inst{1}\orcidID{0009-0008-9660-6117} \and Moysis Symeonides\inst{2}\orcidID{0009-0007-2711-1949} \and Demetris Trihinas\inst{1}\orcidID{0000-0002-9540-7342}
}
\authorrunning{P. Michael et al.}
%
\institute{
Department of Computer Science, University of Nicosia, Cyprus
\email{michael.p15@live.unic.ac.cy, trihinas.d@unic.ac.cy}
\and
Department of Computer Science, University of Cyprus, Cyprus
\email{msymeo03@ucy.ac.cy}
}

\maketitle              
\begin{abstract}
Time Series Foundation Models (TSFMs) promise a paradigm shift toward zero-shot forecasting by eliminating task-specific training. However, existing works often overlook trade-offs between predictive accuracy and probabilistic calibration. This paper presents a benchmark study of six TSFMs evaluated on energy, traffic, and financial datasets. We contrast their performance against statistical baselines and a supervised DL model. The study reveals that while TSFMs outperform statistical methods and supervised models, they are subject to a fundamental trade-off between point accuracy and probabilistic reliability. Specifically, xLSTM architectures provide robust probabilistic calibration across horizons. In contrast, patch-based transformers offer competitive accuracy but face calibration issues at long horizons, while transformer-based models exhibit context saturation points for optimal zero-shot reasoning. These findings offer evidence-based guidance for balancing generalization and uncertainty quantification in real-world deployments.
    

\end{abstract}
\section{Introduction}
\label{sec:intro}
Time series forecasting is important for decision-making in critical domains, including energy grid management, supply chain logistics, and traffic planning. Although statistical models are common, the focus has shifted toward task-specific DL models setting new benchmarks for supervised forecasting~\cite{Nie2023a}. However, these models require extensive historical data, considerable compute for optimization, and frequent retraining under new distributions or domains~\cite{Zhou2021}.

The success of large language models (LLMs) has inspired a new wave of forecasting research known as Time Series Foundation Models (TSFMs). TSFMs are neural networks pre-trained on thousands of time series to enable zero-shot forecasting~\cite{Das2024}. 
This approach bypasses task-specific training, enabling direct deployment on unseen data through mechanisms such as token patching~\cite{liu2026moirai20timeseries}, numerical quantization~\cite{ansari2025chronos2univariateuniversalforecasting}, and generative flow matching~\cite{liu2025sundial}. Consequently, TSFMs offer a practical alternative for cold-start scenarios, such as forecasting demand for new retail products or power generation for newly installed renewable microgrids where historical data is limited~\cite{jin2024timellm}. From a data engineering perspective, TSFMs enable a model-as-a-service paradigm, integrating directly into data pipelines and reducing the complexity of maintaining localized ML/AI workflows.

However, current TSFM evaluations are sparse and fragmented. Recent benchmarks show that while TSFMs are competitive, they do not exceed finely tuned task-specific baselines~\cite{Meyer2025} and under-perform statistical methods for irregular time series~\cite{Toner2025}. Moreover, with TSFMs trained on public data, their performance can be artificially inflated on standard benchmarks. Another critical gap in the literature is the emphasis on point accuracy, neglecting probabilistic calibration. Our work bridges the gap between point accuracy and probabilistic reliability by benchmarking trade-offs among recurrent, generative, and transformer TSFMs, and evaluating them on unseen data to verify their generalization.

The contributions of this work are summarized as follows:
\begin{itemize}
    \item A comprehensive, reproducible, and open-source\footnote{~\url{https://github.com/unic-ailab/TSFM-Eval}} benchmark study on zero-shot time series forecasting, where we evaluate 6 TSFMs (Chronos-2, TiRex, Moirai-2.0, Sundial, TimesFM, and Toto) against statistical baselines and a state-of-the-art supervised DL architecture (PatchTST).
    \item We contrast model performance across diverse data representations 
    and utilize newly collected datasets to ensure zero-shot forecasting on data unseen by TSFMs during their training to assess model robustness.
    \item The evaluation extends beyond point-error metrics, by using Interval Coverage Error (ICE) and Interval Mean Absolute Error (IMAE) to quantify the reliability of the uncertainty estimates from the TSFMs.
\end{itemize}

The rest of this paper is organized as follows: Section 2 reviews related work, Section 3 describes datasets and experimental setup, Section 4 presents the experiments and key takeaways, and Section 5 concludes and outlines future work.

\section{Related Work}
\label{sec:rw}
Prior to foundation models, the state-of-the-art was dominated by task-specific DL architectures trained on target datasets. The Transformer 
advanced this generation through self-attention, enabling models to capture long-range dependencies, with early adaptations such as Informer reducing the quadratic complexity of attention to handle longer contexts~\cite{Zhou2021}. The paradigm matured with PatchTST~\cite{Nie2023a}, which segments time steps into discrete patches and processes variables independently, reducing memory overhead and setting a new supervised benchmark. Despite high accuracy, these models are constrained by large volumes of domain-specific data and significant optimization overhead, limiting their utility in zero-shot or cold-start scenarios~\cite{jin2024timellm}.

TSFMs overcome the limitations of task-specific training by pre-training on large, diverse datasets and using zero-shot inference to generalize across unseen distributions and domains. A key challenge is translating continuous numerical data into a format compatible with architectures designed for natural language. Chronos-2~\cite{ansari2025chronos2univariateuniversalforecasting} applies strict quantization, mapping continuous values into a discrete token vocabulary and treating forecasting as language modeling. Moirai-2.0~\cite{liu2026moirai20timeseries} and TimesFM~\cite{Das2024} instead adopt patch-based, decoder-only autoregressive architectures, where Moirai-2.0 predicts multiple future patches simultaneously via multi-token prediction, and TimesFM lets the output patch length exceed the input patch length for long-horizon generation. Toto~\cite{cohentime}, a transformer-based TSFM embeds a Student-T mixture output head to capture heavy-tailed distributions common in time series observability data.

While attention transformers dominate, recent work proposes generative and state-tracking architectures to better model predictive uncertainty and temporal continuity. Sundial~\cite{liu2025sundial} approaches forecasting through generative modeling, using continuous tokenization and a flow-matching objective to sample multiple future trajectories and construct probabilistic bounds without a fixed parametric distribution, making it adaptable to volatile data. In contrast, TiRex~\cite{auer2025tirex} abandons attention in favor of an extended LSTM (xLSTM) architecture, where its recurrent backbone explicitly tracks hidden states over time, maintaining continuous internal memory for consistent long-range predictions.

\begin{figure}[t!]
    \centering
    \includegraphics[width=\linewidth]{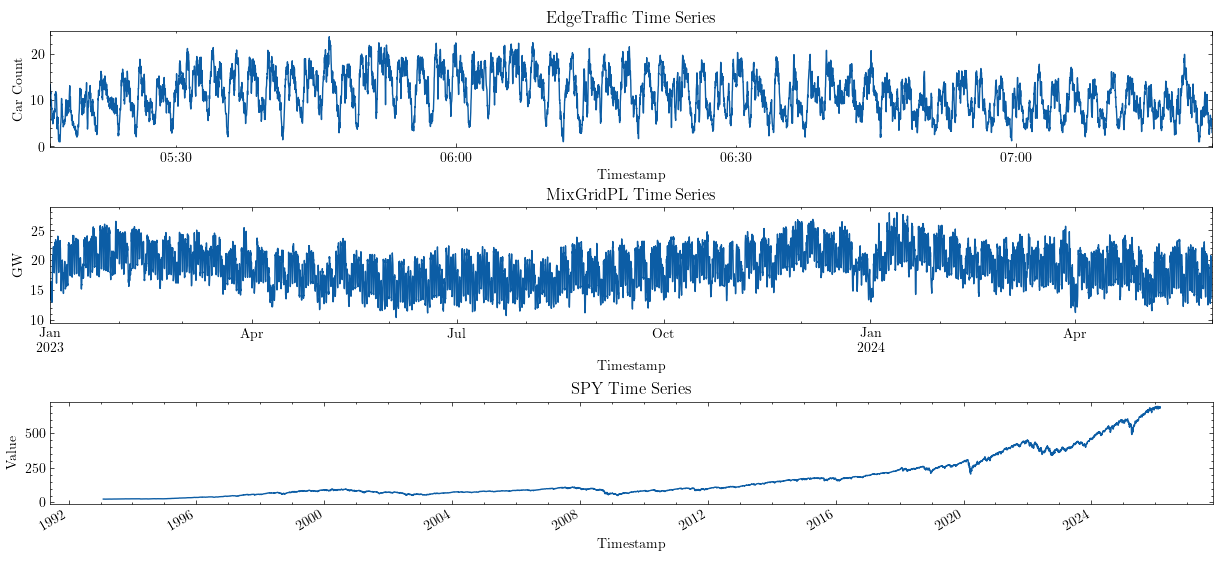}
    \caption{High-Level Overview of the Datasets Embraced for the Benchmarking Study}
    \label{fig:trials}
\end{figure}

\section{Experimental Methodology}
\label{sec:methodology}

\subsection{Datasets}
\label{sec:datasets}
We evaluate zero-shot generalization across three modalities~(Fig.~\ref{fig:trials}): discrete traffic counts (EdgeTraffic), continuous energy load (MixGridPL), and stochastic financial indices (SPY). These span a wide range of characteristics, including high versus low volatility, periodic versus stationary behavior, and temporal granularity. EdgeTraffic and MixGridPL are recently curated data outside known training corpora, isolating true zero-shot reasoning from data leakage.

The \textbf{EdgeTraffic} dataset~\cite{EdgeTraffic} provides high-frequency vehicle counts from a road intersection in Iasi, Romania (7478 observations at 1s resolution over 2h window). The data (Fig.~\ref{fig:trials}a) exhibits high volatility, periodic patterns, and a discrete modality that is challenging for continuous-value foundation models.

The \textbf{MixGridPL} dataset~\cite{CarbonOracle} tracks aggregate electricity for Poland's national power grid (12407 observations at 1h resolution) over 17 months in 2024-25 (Fig.~\ref{fig:trials}b). A continuous, highly seasonal series whose recent collection ensures its temporal patterns were unavailable during TSFM development.

The Yahoo! \textbf{SPY} dataset consists of 8324 daily closing prices for SPDR S\&P 500 ETF from  1993 to 2025 (Figure~\ref{fig:trials}c). SPY is non-stationary with a long-term positive trend but significant short-term volatility, and percentage shifts are evaluated to gauge model responsiveness to economic trends.

\begin{table}[t!]
\centering
\caption{Experimental configurations for varied context ($C$) and horizon ($H$)}
\label{table:sc-configs}
\begin{tabular}{@{}llll@{}}
\toprule
Dataset & Scenario & Fixed Value & Varying Range \\ \midrule
\textbf{EdgeTraffic} & FixedC & $C = 900$ & $H \in \{10, 60,
 300, 600, 900, 1,200, 1,800\}$ \\
 & FixedH & $H = 60$ & $C \in \{120, 300, 600, 900, 1,200, 1,800, 2,400\}$ \\ \midrule
\textbf{MixGridPL} & FixedC & $C = 240$ & $H \in \{12, 24, 72, 144, 288, 576, 672\}$ \\
 & FixedH & $H = 24$ & $C \in \{96, 120, 168, 240, 336, 480, 600\}$ \\ \midrule
\textbf{SPY} & FixedC & $C = 60$ & $H \in \{1, 5, 21, 42, 63, 126, 252\}$ \\
 & FixedH & $H = 5$ & $C \in \{20, 40, 60, 126, 252, 378, 504\}$ \\ \bottomrule
\end{tabular}
\end{table}

\subsection{Problem Description}
\label{sec:problem}
We define a univariate time series as an ordered sequence of observations $\mathcal{X} = \{x_1, x_2, \dots, x_N\}$. For any given reference time $T$, which represents the current moment in the series, the goal of zero-shot forecasting is to predict a future sequence of observations $\hat{\mathcal{X}}_{fut} = \{\hat{x}_{T+1}, \dots, \hat{x}_{T+H}\}$ for a defined horizon $H$ by utilizing a historical context of length $C$ ($x_{T-C+1}, \dots, x_T$) without performing any gradient updates or fine-tuning on the target dataset. 

To evaluate this goal across our experimental scenarios, we employ a non-overlapping rolling window evaluation scheme where each instance is constructed from a segment of length $L = C + 2H$. Each segment is divided into three parts: a context window of length $C$ and two consecutive horizon windows of length $H$. The first horizon window provides ground-truth data to fairly train the supervised DL baseline (PatchTST, Section 3.4), while for the zero-shot models it is merged with the context to form an extended input of $C + H$. The second horizon window remains the consistent inference target across all models, ensuring all models are evaluated on the exact same unseen data. Subsequent segments are produced by sliding the start index forward by $H$, continuing while
a full segment of length $C + 2H$ fits within the series.

\subsection{Experiment Scenarios}
\label{sec:scenarios}
To evaluate model performance, two parameters are varied for experimentation: the historical context $C$ and prediction length $H$, as shown in Table~\ref{table:sc-configs}.

\noindent \textbf{-- Scenario 1 - Fixed Context (FixedC)}. For each dataset, we fix a representative
 context $C$ while varying the forecast horizon $H$ from short to long-range. This assesses how accuracy scales as the prediction distance increases.

\noindent \textbf{-- Scenario 2 - Fixed Horizon (FixedH).} We keep the horizon $H$ constant and vary the context length $C$. This identifies the point of ``context saturation'', beyond which additional history no longer improves zero-shot reasoning.


\subsection{Benchmark Baselines and TSFMs}
\label{sec:models}

We evaluate three statistical baselines: (i) a fixed-order \textbf{ARIMA} model with local autocorrelation and non-stationarity; (ii) a \textbf{Running Average (RA)} adopting the context mean over the prediction horizon; and (iii) a \textbf{Random Walk with Drift (RWD)}, setting each future value to the previous plus a drift estimated from historical variance. These are configured in their best settings across the datasets. We also use a supervised DL baseline, \textbf{PatchTST}, configured for architectural and computational parity with the TSFMs (single encoder layer and attention head, hidden dimension of 32, and 0\% dropout), to achieve inference latency comparable to benchmarked TSFMs. It also uses robust scaling to manage outliers, and is optimized with Multi-Quantile Loss (MQLoss) to estimate 10th, 50th, and 90th percentiles for direct probabilistic comparison.

We evaluate six state-of-the-art TSFMs that span diverse architectures and scales (detailed in Section 2), enabling a comparative analysis of how internal structure influences forecasting accuracy across the evaluated modalities. All adopt optimal pre-trained configurations to ensure standardized zero-shot evaluation and only adjust each model to emit the quantile forecasts required by the calibration metrics in Section~\ref{sec:testbed}. The benchmarked models and their parameter counts are: \textbf{Moirai-2} (11.4M), \textbf{TiREX} (35M), \textbf{Chronos-2} (120M), \textbf{Sundial} (128M), \textbf{Toto} (151M), and \textbf{TimesFM} (200M).

\subsection{Testbed and Evaluation Metrics}
\label{sec:testbed}

Experiments are run in a virtual realm with an Nvidia T4 GPU (15GB VRAM) and 12GB of system RAM. Implementation is based on PyTorch, with all TSFMs originating from Hugging Face in their default settings. Inference is performed without task-specific fine-tuning to evaluate zero-shot
forecasting.

\begin{figure}[t!]
    \centering
    \includegraphics[width=0.94\linewidth, trim=0.4cm 0.3cm 0.4cm 0.0cm, clip]{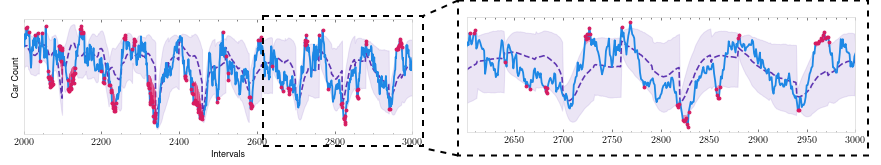}
    \caption{Example TSFM with EdgeTraffic dataset depicting ground truth (blue line), forecast (purple line), confidence interval (shaded area), and interval violations (pink points). Zoomed-in highlight of last 500 datapoints provides a detailed overview.}
    \label{fig:example}
\end{figure}

We employ three metrics for predictive accuracy and calibration. For point accuracy, the Symmetric Mean Absolute Percentage Error (sMAPE, eq. 1) provides a percentage error measure robust to scale variations normalizing the absolute error by the mean magnitude of the observed and predicted values. For probabilistic reliability, we use Interval Coverage Error (ICE) and Interval Mean Absolute Error (IMAE) to evaluate prediction intervals. We use an 80\% prediction interval with quantiles $\alpha=0.1$ and $\beta=0.9$. Fig.~\ref{fig:example} depicts this setting, with the ground truth, point predictions, and shaded intervals between the quantile bounds. ICE in eq. (2) measures calibration error by comparing the observed miscoverage, the fraction of points outside the shaded band, with the nominal
miscoverage of $0.20$. A value of zero indicates perfect calibration. IMAE in eq. (3) complements ICE by measuring the magnitude of these violations, the vertical distance between each out-of-interval point and the nearest interval bound in Fig.~\ref{fig:example}. Thus, ICE captures how often the interval fails, while IMAE captures how severe these failures are. A model may therefore achieve low ICE but high IMAE if only a few observations fall outside the interval, but by a large margin.
\begin{align}
\text{sMAPE} &= \frac{100}{H} \sum_{h=1}^{H} 
\begin{cases} 
0, & \text{if } |x_{T+h}| + |\hat{x}_{T+h}| = 0 \\
\dfrac{2 \, | x_{T+h} - \hat{x}_{T+h} |}{|x_{T+h}| + |\hat{x}_{T+h}|}, & \text{otherwise}
\end{cases}\\
\text{ICE}_{\alpha,\beta} &= \left| \frac{1}{H} \sum_{h=1}^{H} (\mathbbm{1}_{\{ x_{T+h} < \hat{x}^{\alpha}_{T+h} \vee x_{T+h} > \hat{x}^{\beta}_{T+h} \}}) - (1 - (\beta - \alpha)) \right| \\
\text{IMAE}_{\alpha,\beta} &= \frac{1}{H} \sum_{h=1}^{H} \left[ \max(0, \hat{x}^{\alpha}_{T+h} - x_{T+h}) + \max(0, x_{T+h} - \hat{x}^{\beta}_{T+h}) \right]
\end{align}

Here, $\hat{x}^{\alpha}_{T+h}$ and $\hat{x}^{\beta}_{T+h}$ represent the predicted $\alpha$- and $\beta$-quantiles at horizon step $h$, respectively. In turn, $T$ denotes the reference time, while $h = 1, \dots, H$ indexes the forecast horizon of length $H$. The variable $x_{T+h}$ represents the ground-truth, and $\hat{x}_{T+h}$ denotes the corresponding median point prediction.

\section{Evaluation}
\begin{figure}[t]
    \centering
    \includegraphics[width=\linewidth, trim=0.2cm 0.3cm 0.3cm 0.3cm, clip]{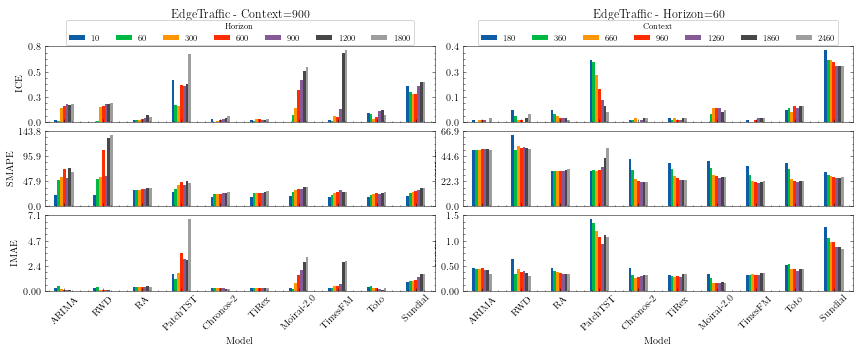}
    \caption{Overview of Benchmark Results for the EdgeTraffic Dataset}
    \label{fig:EdgeTraffic}
\end{figure}

\begin{figure}[t]
    \centering
    \includegraphics[width=\linewidth, trim=0.22cm 0.45cm 0.15cm 0.35cm, clip]{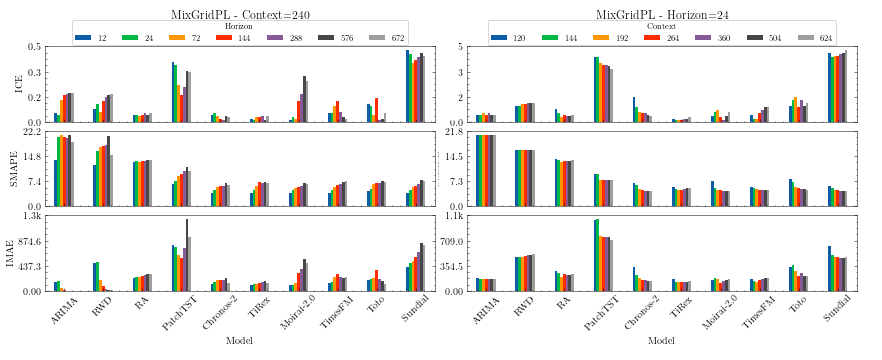}
    \caption{Overview of Benchmark Results for the MixGridPL Dataset}
    \label{fig:MixGridPL}
\end{figure}

\begin{figure}[t]
    \centering
    \includegraphics[width=\linewidth]{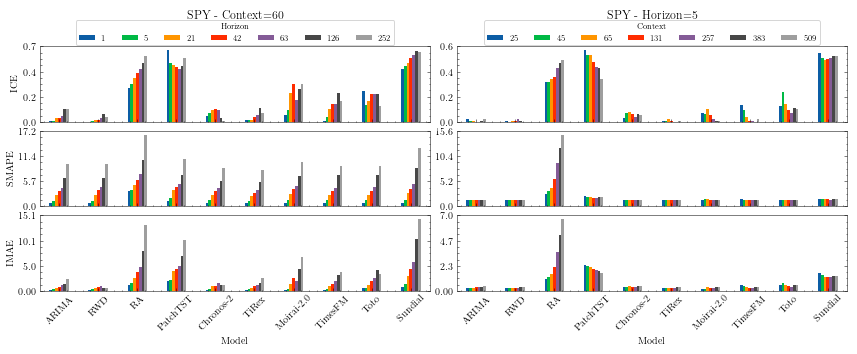}
    \caption{Overview of Benchmark Results for the SPY Dataset}
    \label{fig:SPY}
\end{figure}

This section examines the benchmark results for predictive accuracy and probabilistic calibration, visualized per dataset in Figures~\ref{fig:EdgeTraffic}-\ref{fig:SPY}. For brevity, findings are presented as plots. Full tabular data and reproduction instructions are openly available in the benchmark repository.

\subsection{Predictive Accuracy Evaluation of Forecasting Models}

The accuracy results show that TSFMs perform strongly across diverse settings, though their advantage depends on the structure of the underlying series. In volatile realms they remain more robust than traditional statistical methods, particularly as the forecast horizon extends.

For the discrete EdgeTraffic dataset and FixedC scenario, TimesFM and Chronos-2 achieve the best short-horizon sMAPE ($17\%$) at $H=10$, surpassing ARIMA ($22\%$) and RA ($32\%$). The gap widens at long horizons: at $H=1800$ Chronos-2 remains the most stable ($28\%$), while ARIMA, RA, and RWD exceed $65\%$, $35\%$, and $136\%$. Under FixedH, TSFMs show non-monotonic sensitivity to context. TimesFM improves from $36\%$ at $C=180$ to $21\%$ at $C=1260$, then marginally degrades at $C=2460$ ($22.5\%$), indicating that excessive history can introduce noise rather than improve accuracy. In contrast, supervised PatchTST collapses as context grows, rising from $31.5\%$ at $C=180$ to $52\%$ at $C=2460$. This exposes a key limitation of task-specific architectures, where they lack the ``context reasoning'' of TSFMs and overfit as input dimensionality grows.

Gains are more pronounced in the MixGridPL dataset, where TSFMs substantially outperform statistical baselines for all scenarios. For FixedC, Chronos-2, TiRex, and Sundial reach $3.92\%$ to $3.95\%$ at $H=12$, versus $14\%$ for ARIMA and $13\%$ for RA; at $H=672$, Chronos-2 and Moirai-2.0 stay at $6.4\%$ and $6.6\%$ while ARIMA reaches $19\%$. The FixedH scenario confirms that TSFMs exploit longer contexts for seasonal data, with Chronos-2 improving from $6.70\%$ at $C=120$ to $4.4\%$ at $C=624$ and Moirai-2.0 following a similar trend.


For SPY, short-horizon results are nearly identical across methods (at $H=1$, RWD $0.79\%$, TiRex $0.80\%$, Chronos-2 $0.82\%$), but TSFMs become competitive at long horizons, where TiRex ($8.34\%$) and Chronos-2 ($8.64\%$) at $H=252$ improve over ARIMA ($9.68\%$) and RWD ($9.74\%$). Under FixedH ($H=5$), most TSFMs remain stable as context grows (TiRex from $1.40\%$ at $C=25$ to $1.26\%$ at $C=509$), whereas the RA worsens from $2.50\%$ to $14.83\%$, confirming that a constant-mean approach is unsuitable for trending financial data.

\noindent \textbf{Key Takeaway:} \textit{TSFMs consistently match or exceed traditional baselines, with the largest gains in seasonal and structurally rich series where long-context reasoning provides a clear zero-shot advantage.}

\subsection{Probabilistic Reliability and Interval Calibration}

Several of the TSFMs maintain low sMAPE at extended horizons, but their uncertainty estimates deteriorate, leading to significant miscalibration, as shown by their ICE and IMAE scores. 

For EdgeTraffic (FixedC), several transformer models suffer a \emph{calibration collapse} as the horizon extends. TimesFM and Moirai rise from near-perfect calibration at $H=10$ to ICE values of $0.73$ and $0.56$ at $H=1800$, indicating severe over-confidence and a failure to capture observed traffic peaks where their IMAE also scales poorly (reaching $2.8$ and $3.2$). This denotes that missed values are missed by a wide margin. Toto is the exception among transformers, holding an ICE of $0.07$ at $H=1800$, which suggests its Student-T mixture head captures heavy-tailed distributions more robustly than the standard density heads of other patch-based models. The best performers are Chronos-2 and TiRex, both keeping ICE below $0.06$ across all horizons and their slightly higher IMAE (averaging ${\sim}0.28$) reflects a conservative profile that is preferable when missing a traffic burst is costlier than overestimating variance. PatchTST performs worst, with an ICE of $0.7$ and IMAE of $6.8$ at $H=1800$, showing that task-specific training fails to generalize uncertainty to long-range and out-of-distribution shifts.

For MixGridPL, high seasonality yields more stable behavior, although differences persist. TiRex has an ICE between $0.01$ and $0.04$ across all scenarios. Under FixedH, Chronos-2 improves its calibration as context grows, with ICE dropping from $0.17$ ($C=120$) to $0.04$ ($C=624$). In contrast, Sundial and PatchTST are over-confident regardless of context (ICE ${\sim}0.45$). Moirai achieves good short-horizon calibration (ICE $0.01$ at $H=12$), but long-range reliability degrades sharply, with IMAE rising from $109$ to $479$, indicating patch-based transformers struggle to propagate uncertainty through long autoregressive chains.

For SPY, the stochastic signal exposes narrow interval risk. Sundial and PatchTST exceed a  $0.50$ ICE, missing half of the price movements they target, while TimesFM and Moirai-2.0 show similar decay as the horizon extends. TiRex and Chronos-2 are the most robust. Under FixedH, as context grows to $C=509$, TiRex reaches a near-perfect ICE of $0.01$ with a low IMAE ($0.34$), confirming its intervals are safe without being excessively loose. This balance makes recurrent (TiRex) and quantized-encoder (Chronos-2) architectures more suitable for risk-sensitive financial applications than purely generative or patch-based ones.

\noindent \textbf{Key Takeaway:} \textit{Point accuracy does not guarantee probabilistic. TimesFM and Moirai often lead in sMAPE yet become over-confident at long horizons, whereas Chronos and TiRex provide calibrated uncertainty across modalities.}

\section{Conclusion and Future Directions}
\label{sec:conclusion}

This paper presented a benchmark of zero-shot TSFMs across different data modalities. TSFMs emerge as a strong alternative to statistical and lightweight supervised baselines, especially in seasonal settings where longer context windows improve performance. However, point accuracy alone is insufficient to assess real-world deployment where forecasting quality depends on a trade-off between predictive accuracy and uncertainty calibration. TSFM architecture drives this trade-off. The xLSTM-based TiRex gives the most consistent, conservative calibration across modalities, staying stable as the horizon extends, whereas patch-based transformers such as Moirai-2.0 and TimesFM often lead on accuracy yet suffer a calibration collapse, becoming over-confident at long horizons. Chronos-2 is the most balanced, pairing competitive accuracy with well-calibrated intervals. Success is also signal-dependent, where seasonal data (MixGridPL) is most favorable, discrete high-frequency traces (EdgeTraffic) challenge calibration, and stochastic
financial series (SPY) with short-term volatility remain hardest. TSFM selection should thus weigh forecast horizon, signal volatility, and the cost of under-estimating uncertainty, not sMAPE alone.

Future work includes extending the benchmark to multivariate forecasting and to robustness under missing data, irregular sampling, and concept drift. A dedicated study of computational efficiency is left to forthcoming work, alongside calibration-aware forecasting and adaptive model-selection strategies.


%
%
%
 \bibliographystyle{splncs04}
 \bibliography{mybib}

\end{document}